\documentclass{article}
\usepackage{spconf,amsmath,graphicx}

\usepackage{url}
\usepackage{multirow}
\usepackage{caption}
\usepackage{rotating}

\usepackage{lipsum}

\newcommand\blfootnote[1]{%
  \begingroup
  \renewcommand\thefootnote{}\footnote{#1}%
  \addtocounter{footnote}{-1}%
  \endgroup
}

\title{MambaNet: A Hybrid Neural Network for Predicting the NBA Playoffs}
\name{Reza Khanmohammadi$^{\dagger}$ \; Sari Saba-Sadiya$^{\dagger}$ \; Sina Esfandiarpour$^{\mathsection}$ \; \textit{Tuka Alhanai}$^{\ddagger}$ \; \textit{Mohammad M. Ghassemi}$^{\dagger}$}

\address{$^{\dagger}$Computer Science and Engineering Department, Michigan State University \\
$^{\mathsection}$Basketball Analyst, InStat Sports Performance Analysis Company \\
$^{\ddagger}$Department of Electrical \& Computer Engineering, New York Univesity Abu Dhabi}

\begin{document}
%
\maketitle
\begin{abstract}




\noindent In this paper, we present Mambanet: a hybrid neural network for predicting the outcomes of Basketball games. Contrary to other studies, which focus primarily on season games, this study investigates playoff games. MambaNet is a hybrid neural network architecture that processes a time series of teams' and players' game statistics and generates the probability of a team winning or losing an NBA playoff match. In our approach, we utilize Feature Imitating Networks to provide latent signal-processing feature representations of game statistics to further process with convolutional, recurrent, and dense neural layers. Three experiments using six different datasets are conducted to evaluate the performance and generalizability of our architecture against a wide range of previous studies. Our final method successfully predicted the AUC from 0.72 to 0.82, beating the best-performing baseline models by a considerable margin.
\end{abstract}
\section{Introduction}
\vspace{-0.5cm}
\label{sec:intro}

\noindent \blfootnote{$^{}$ T Alhanai and MM Ghassemi contributed equally to this work and should be considered shared senior authors.}

\noindent Sporting events are a popular source of entertainment, with immense interest from the general public. Sports analysts, coaching staff, franchises, and fans alike all seek to forecast winners and losers in upcoming sports match-ups based on previous records. The interest in predicting sporting outcomes is particularly pronounced for professional team sport leagues including the Major League Baseball (MLB), the National Football League (NFL), the National Hockey League (NHL), and the  National Basketball Association (NBA); postseason plays in these leagues, namely the playoffs, are are of greater interest than games in the regular season because teams compete directly for prestigious championships titles. 

The development of statistical models to robustly predict the outcome of playoff games from year-to-year is a challenging machine learning task because of the plethura of individual, team and extenral factors that all-together confound the propensity of a given team to win a given game in a give year. In this work, we develop \textbf{MambaNet}: a large hybrid neural network for predicting the outcome of a basketball match during the playoffs. There are five main differences between our work and previous studies: (1) we use a combination of both player and team statistics;(2) we account for the evolution in player and team statistics over time using a signal processing approach; (3)  we utilize Feature Imitating Networks (FINs) \cite{sari} to embed feature representations into the network; (4) we predict the outcome of playoff results, as opposed to season games; and (5) we test the generalizability of our model across two distinct national basketball leagues. To assess the value of our proposed approach, we performed three experiments that compare MambaNet's to previously-proposed machine learning algorithms using NBA and Iranian Super League data. 


\begin{figure*}[t]
\includegraphics[width=\textwidth]{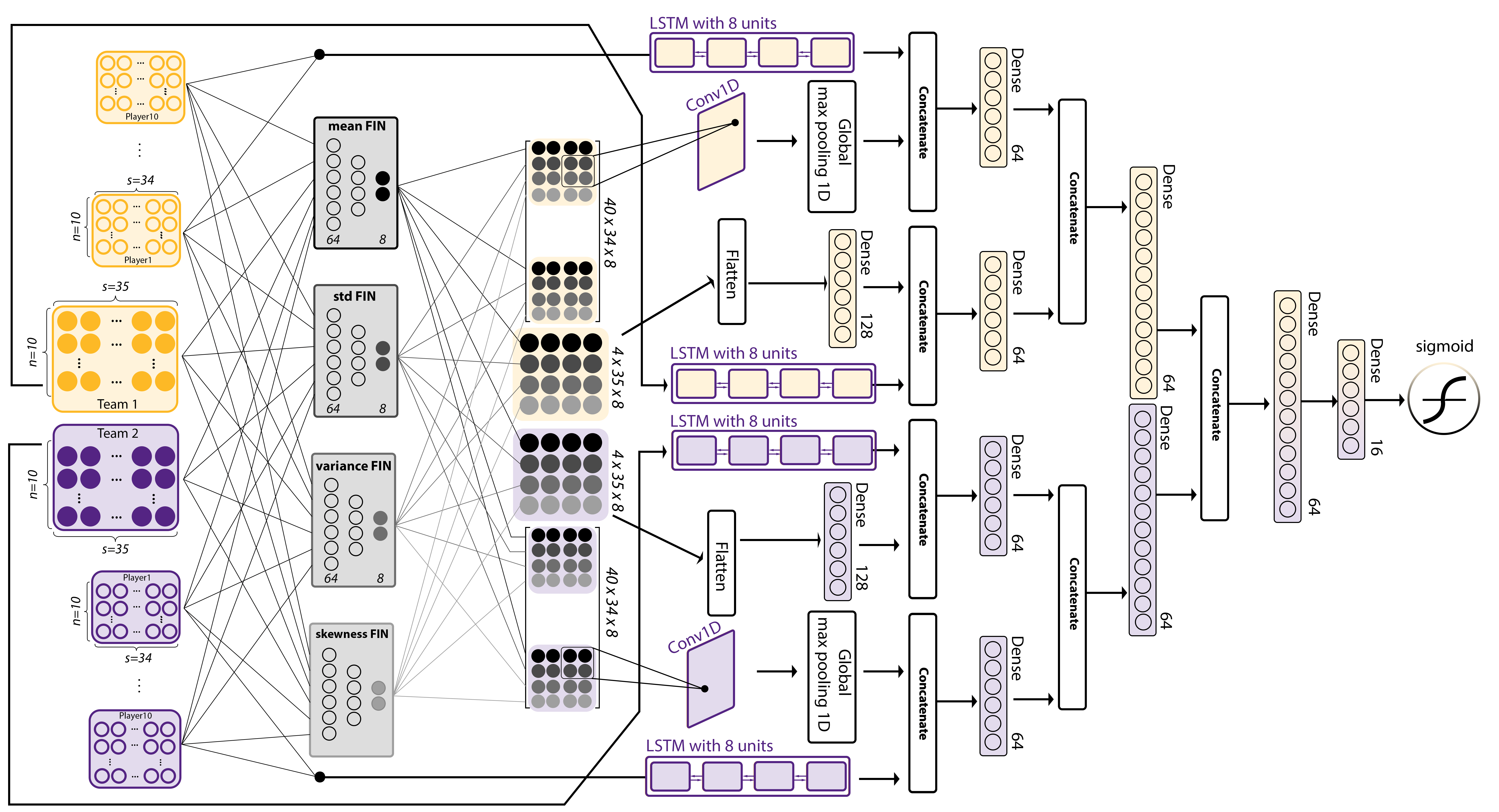}
\caption{\label{fig:mambanet}An overview of MambaNet's architecture. First, the home (column 1, yellow boxes) and away (column 1, purple boxes) teams' stats and the two teams' players' stats are fed to the network. Next, four FINs are utilized to represent the input stats' signal features, which contain trainable (column 1, dark circles) and non-trainable (column 2, light circles) layers. These representations are further processed with convolutional and Dense layers. Raw time-domain signal features are also extracted from input stats using LSTM networks. Finally, the aforementioned features are incorporated to make the final prediction.}
\centering
\end{figure*}

\bgroup
\def\arraystretch{0.7}
\begin{table*}[t]
\centering
\footnotesize
\captionsetup{font=footnotesize}
\begin{tabular}{cl|cl|cl}
\hline
\# & Statistics                      & \# & Statistics                  & \# & Statistics                         \\ \hline
1  & Minutes Played                    & 2  & Field Goal                   & 3  & Field Goal Attempts               \\
4  & Field Goal Percentage         &  5  & 3-Point Field Goal           & 6  & 3-Point Field Goal Attempts       \\
7  & 3-Point Field Goal Percentage & 8 & Free Throw                  & 9  & Free Throw Attempts            \\
10 & Free Throw Percentage         & 11 & Offensive Rebound           & 12 & Defensive Rebound                  \\
13 & Total Rebound                   & 14 & Assists                    & 15 & Steals                            \\
16 & Blocks                          & 17 & Turnover             & 18 & Personal Fouls       \\
19 & Points                         & 20 & True Shooting Percentage &  21 & Effective Field Gaol Percentage \\
22 & 3-Point Attempt Rate          & 23 & Free Throw Attempt Rate    & 24 & Offensive Rebound Percentage \\
25 & Defensive Rebound Percentage & 26 & Total Rebound Percentage  & 27 & Assist Percentage             \\
28 & Steal Percentage              & 29 & Block Percentage         & 30 & Turnover Percentage              \\
31 & Usage Percentage               & 32 & Offensive Rating       & 33 & Defensive Rating                  \\
34 & Winning Percentage (Team-only)  & 35 & Elo Rating (Team-only)   &  36 & Plus/Minus (Player-only)          
\end{tabular}
\caption{\label{tab:stats}A description of game statistics used in this work. Except for the last three features, the rest (1 to 33) are shared statistics in representing both teams and players.  (Abr: Abbreviation, \#: feature number)}
\end{table*}
\egroup

\section{Related Work}
\label{sec:pagestyle}
The NBA is the most popular contemporary basketball league \cite{Ulas2021ExaminationON,kawashiri2020societal}. Several previous studies have examined the impact of different game statistics on a team's propensity to win or lose a game \cite{ijerph17165722,seconndassist}. More specifically, previous studies have identified teams' defensive rebounds, field goal percentage, and assists as crucial contributing factors to succeeding in a basketball game \cite{reb}; for machine learning workflows, these game attributes may be used as valuable input features to predict the outcome of a given basketball game \cite{mlbs,BUNKER201927}.

Probabilistic models to predict the outcome of basketball games have been proposed by several previous studies. Jain and Kaur \cite{kaur} developed a Support Vector Machine (SVM) and a Hybrid Fuzzy-SVM model (HFSVM) and reported 86.21\% and 88.26\% accuracy in predicting the outcome of basketball games. More recently, Houde \cite{houde} experimented with SVM, Gaussian Naive Bayes (GNB), Random Forest (RF) Classifier, K Neighbors Classifier (KNN), Logistic Regression (LR), and XGBoost Classifier (XGB) over fifteen game statistics across the last ten games of both home and away teams. They also experimented over a more extended period of NBA season data, starting from 2018 to 2021, and reported 65.1\% accuracy in winners/losers classifications. In contrast to Kaur and Houde, that addressed game outcome prediction as a binary classification task, Chen et al \cite{chen} identified the winner/loser by predicting their exact final game scores. They used a data mining approach, experimenting with 13 NBA game statistics from the 2018-2019 season. After feature selection, this number shrank to 6 critical basketball statistics for predicting the outcome. In terms of classifiers, the authors experimented with KNN, XGB, Stochastic Gradient Boosting (SGB), Multivariate Adaptive Regression Splines (MARS), and Extreme Learning Machine (ELM) to train and classify the winner of NBA matchups. The authors also studied the effect of different game-lag values (from 1 to 6) on the success of their utilized classifiers and indicated that 4 was found to perform best on their feature set. 

Fewer studies have used Neural Networks to predict the outcome of basketball games; this is mostly due to challenges of over-fitting in the presence of (relatively) small basketball training datasets. Thabtah et al \cite{thab} trained Artificial Neural Networks (ANN) on a wide span of data where they extracted 20 team stats per NBA matchup played from 1980 to 2017. Their model obtained 83\% accuracy in predicting NBA game outcomes; they also demonstrated the significance of three-point percentage, free throws made, and total rebounds as features that enhanced their model's accuracy rate. 

\section{METHODS}
\label{sec:pagestyle}
\noindent \textbf{Baseline approach}: A majority of the existing studies use a similar methodological approach: For each team (home and away), a set of $s$ game statistics (the features) are extracted over $n$ previous games (the game-lag value \cite{chen}) forming an $n \times s$ matrix. Then, the mean of each stat is calculated across the n games, resulting in a $1 \times s$ feature vector for each team. The two feature vectors and concatenated yielding a $1 \times 2s$ vector for each unique matchup between a given pair of teams. Finally, this results in a $trainSize \times 2s$ matrix which is used to train classification model (each experiment will report the train/set set size in more detail). Alternatively, the label of each sample indicates whether the home team won ($y=1$) or lost the game ($y=0$).

\noindent \textbf{FIN Training}: Our method follows the same steps as the baseline approaches, but with one critical difference: instead of calculating the mean of features across the $n$ last games using the mean equation, we feed the entire $n \times s$ matrix to a pretrained mean FIN and stack hidden layers on top of it (hereafter, this FIN-based deep feedforward architecture is referred to as FINDFF) to perform binary classification; In addition to the mean feature, we also imitate standard deviation, variance, and skewness. 

All FINs are trained using the same neural architecture: A sequence of dense layers with 64, 32, 16, 8, and 4 units are stacked, respectively, before connecting to a single-unit sigmoid layer. The activation function of is ReLU, for the first two hidden layers and the rest are Linear. Each model is trained in a regression setting by using 100,000 randomly generated signals as the training set and handcrafted feature values for each signal as the training labels. Then, we freeze the first three layers, finetune the fourth layer, and remove the remaining two layers before integrating them within the larger network structure of the MambaNet network.

\noindent \textbf{Mambanet}: In Figure \ref{fig:mambanet}, we provide an illustration of Mambanet - our proposed approach. The complete set of player and team statistics used in this study can also be found in Table \ref{tab:stats}. The input to the network is an $10 \times 35$ stats matrix which are passed to both the pretrained FINs as well as LSTM layers to extract a \textit{team}'s statistics' sequential features. For each team, we also extract the a stats matrix ($n=10$, $s=34$) for each of its roster's top ten \textit{players} and pass them to the same FINs and LSTM layers. Next, we flatten teams' signal feature representations and feed them to dense layers, whereas for players, we stack them and feed them to 1D convolutional layers. Finally, all latent representations of a team and its ten players are concatenated in the network before connecting them to the last sigmoid layer.  

\section{Experiments \& Results}
\label{sec:typestyle}

We performed three experiments to assess the performance of our proposed method. To demonstrate the advantage of leveraging FINs in deep neural networks, we first compare the performance of FINDFF against a diverse set of other basketball game outcome prediction models trained using NBA data. For the second experiment, these models are tested for generalization across unseen basketball playoff games from the Iranian Super League data. Finally, we assess the performance of Mamabanet for accurate playoff outcome prediction. In all three experiments, the Area Under the ROC Curve (AUC) was used as our primary evaluation metric. 

\bgroup
\def\arraystretch{1}
\begin{table}[t]
\centering
\footnotesize
\captionsetup{font=footnotesize}
\begin{tabular}{|c|c|c|c|c|c|c|c|}
\hline
\multirow{2}{*}{Ref} &
  \multirow{2}{*}{FC} &
  \multirow{2}{*}{Alg} &
  \multicolumn{5}{c|}{AUC} \\ \cline{4-8}
   & & & 
  \multicolumn{1}{c|}{17-18} &
  \multicolumn{1}{c|}{18-19} &
  \multicolumn{1}{c|}{19-20} &
  \multicolumn{1}{c|}{20-21} &
  21-22 \\ \hline \hline
{\cite{kaur}} &
  33 &
  SVM &
  \multicolumn{1}{c|}{0.65} &
  \multicolumn{1}{c|}{0.57} &
  \multicolumn{1}{c|}{0.59} &
  \multicolumn{1}{c|}{0.50} &
  0.60 \\ \hline
\textbf{TW} &
  \textbf{33} &
  \textbf{FINDFF} &
  \multicolumn{1}{c|}{\textbf{0.71}} &
  \multicolumn{1}{c|}{\textbf{0.62}} &
  \multicolumn{1}{c|}{\textbf{0.71}} &
  \multicolumn{1}{c|}{\textbf{0.55}} &
  \textbf{0.65} \\ \hline \hline
\multirow{6}{*}{\cite{houde}} &
  \multirow{6}{*}{15} &
  GNB &
  \multicolumn{1}{c|}{0.60} &
  \multicolumn{1}{c|}{0.52} &
  \multicolumn{1}{c|}{0.60} &
  \multicolumn{1}{c|}{0.52} &
  0.55 \\ \cline{3-8} 
 &
   &
  RF &
  \multicolumn{1}{c|}{0.62} &
  \multicolumn{1}{c|}{0.62} &
  \multicolumn{1}{c|}{0.60} &
  \multicolumn{1}{c|}{0.58} &
  0.60 \\ \cline{3-8} 
 &
   &
  KNN &
  \multicolumn{1}{c|}{0.49} &
  \multicolumn{1}{c|}{0.53} &
  \multicolumn{1}{c|}{0.63} &
  \multicolumn{1}{c|}{0.60} &
  0.64 \\ \cline{3-8} 
 &
   &
  SVM &
  \multicolumn{1}{c|}{0.55} &
  \multicolumn{1}{c|}{0.53} &
  \multicolumn{1}{c|}{0.64} &
  \multicolumn{1}{c|}{0.51} &
  0.61 \\ \cline{3-8} 
 &
   &
  LR &
  \multicolumn{1}{c|}{0.61} &
  \multicolumn{1}{c|}{0.65} &
  \multicolumn{1}{c|}{0.65} &
  \multicolumn{1}{c|}{0.61} &
  0.66 \\ \cline{3-8} 
 &
   &
  XGB &
  \multicolumn{1}{c|}{0.63} &
  \multicolumn{1}{c|}{0.67} &
  \multicolumn{1}{c|}{0.65} &
  \multicolumn{1}{c|}{0.50} &
  0.59 \\ \hline
\textbf{TW} &
  \textbf{15} &
  \textbf{FINDFF} &
  \multicolumn{1}{c|}{\textbf{0.68}} &
  \multicolumn{1}{c|}{\textbf{0.77}} &
  \multicolumn{1}{c|}{\textbf{0.69}} &
  \multicolumn{1}{c|}{\textbf{0.76}} &
  \textbf{0.70} \\ \hline \hline
{\cite{e18120450}} &
  14 &
  NBAME &
  \multicolumn{1}{c|}{0.51} &
  \multicolumn{1}{c|}{0.53} &
  \multicolumn{1}{c|}{0.53} &
  \multicolumn{1}{c|}{0.57} &
  0.59 \\ \hline
\textbf{TW} &
  \textbf{14} &
  \textbf{FINDFF} &
  \multicolumn{1}{c|}{\textbf{0.62}} &
  \multicolumn{1}{c|}{\textbf{0.59}} &
  \multicolumn{1}{c|}{\textbf{0.64}} &
  \multicolumn{1}{c|}{\textbf{0.60}} &
  \textbf{0.62} \\ \hline \hline
\multirow{5}{*}{\cite{chen}} &
  \multirow{5}{*}{6} &
  ELM &
  \multicolumn{1}{c|}{0.53} &
  \multicolumn{1}{c|}{0.55} &
  \multicolumn{1}{c|}{0.55} &
  \multicolumn{1}{c|}{0.53} &
  0.64 \\ \cline{3-8} 
 &
   &
  KNN &
  \multicolumn{1}{c|}{0.58} &
  \multicolumn{1}{c|}{0.53} &
  \multicolumn{1}{c|}{0.51} &
  \multicolumn{1}{c|}{0.56} &
  0.55 \\ \cline{3-8} 
 &
   &
  XGB &
  \multicolumn{1}{c|}{0.60} &
  \multicolumn{1}{c|}{0.58} &
  \multicolumn{1}{c|}{0.53} &
  \multicolumn{1}{c|}{0.53} &
  0.55 \\ \cline{3-8} 
 &
   &
  MARS &
  \multicolumn{1}{c|}{0.63} &
  \multicolumn{1}{c|}{0.53} &
  \multicolumn{1}{c|}{0.53} &
  \multicolumn{1}{c|}{0.59} &
  0.57 \\ \hline
\textbf{TW} &
  \textbf{6} &
  \textbf{FINDFF} &
  \multicolumn{1}{c|}{\textbf{0.69}} &
  \multicolumn{1}{c|}{\textbf{0.65}} &
  \multicolumn{1}{c|}{\textbf{0.57}} &
  \multicolumn{1}{c|}{\textbf{0.63}} &
  \textbf{0.66} \\ \hline \hline
{\cite{thab}} &
  20 &
  ANN &
  \multicolumn{1}{c|}{0.55} &
  \multicolumn{1}{c|}{0.55} &
  \multicolumn{1}{c|}{0.53} &
  \multicolumn{1}{c|}{0.58} &
  0.53 \\ \hline
\textbf{TW} &
  \textbf{20} &
  \textbf{FINDFF} &
  \multicolumn{1}{c|}{\textbf{0.59}} &
  \multicolumn{1}{c|}{\textbf{0.68}} &
  \multicolumn{1}{c|}{\textbf{0.67}} &
  \multicolumn{1}{c|}{\textbf{0.61}} &
  \textbf{0.65} \\ \hline
\end{tabular}
\caption{\label{tab:pw}A performance comparison between FINDFF and other previously-developed machine learning models on five years of NBA Playoffs, from the 2017-2018 season (17-18) to the 21-22 season. (Ref: Reference, FC: Feature Count, Alg: Algorithm, TW: This Work)}
\end{table}
\egroup

\newcommand{\RNum}[1]{\uppercase\expandafter{\romannumeral #1\relax}}

\subsection{Experiment \RNum{1}}
This experiment aims to determine whether using FINs in conjunction with deep neural networks can enhance playoff outcome prediction. We followed the same machine learning pipeline as previous studies to compare FINDFF. However, we applied a pretrained mean FIN to the $n \times s$ matrix instead of taking the mean directly, providing an almost identical setting when comparing FINDFF with other classic machine learning algorithms. Since the FIN is the only differing component in this setting, its effects can be easily studied. 
\noindent \textbf{Dataset:} All data were gathered from NBA games played from 2017-2018 to 2021-2022 over five seasons. We used each year's season games as training data (1,230, 1,120, 1,060, 1,086, and 1,236 games from 2017-2018 to 2021-2022) and playoff games as testing data (82, 82, 83, 85, and 87 games from 2017-2018 to 2021-2022), leaving us with five different NBA datasets.

\noindent \textbf{Results:} In Table \ref{tab:pw}, we compare FINDFF models with five other methods from the literature using different features (game statistics), game-lag values, and classification algorithms. The FINDFF network successfully outperformed all other methods with a 0.05 to 0.15 AUC margin in every year of NBA data, demonstrating the advantage of the feature imitation technique in game outcome prediction.

\bgroup
\def\arraystretch{1}
\begin{table}[t]
\centering
\footnotesize
\captionsetup{font=footnotesize}
\begin{tabular}{|c|c|c|ccccc|}
\hline
\multirow{2}{*}{Ref} &
  \multirow{2}{*}{FC} &
  \multirow{2}{*}{Alg} &
  \multicolumn{5}{c|}{AUC} \\ \cline{4-8} 
 &
   &
   &
  \multicolumn{1}{c|}{17-18} &
  \multicolumn{1}{c|}{18-19} &
  \multicolumn{1}{c|}{19-20} &
  \multicolumn{1}{c|}{20-21} &
  21-22 \\ \hline \hline
{[}4{]} &
  33 &
  SVM &
  \multicolumn{1}{c|}{0.52} &
  \multicolumn{1}{c|}{0.55} &
  \multicolumn{1}{c|}{0.55} &
  \multicolumn{1}{c|}{0.57} &
  0.55 \\ \hline
\textbf{TW} &
  \textbf{33} &
  \textbf{FINDFF} &
  \multicolumn{1}{c|}{\textbf{0.62}} &
  \multicolumn{1}{c|}{\textbf{0.57}} &
  \multicolumn{1}{c|}{\textbf{0.67}} &
  \multicolumn{1}{c|}{\textbf{0.72}} &
  \textbf{0.59} \\ \hline \hline
\multirow{6}{*}{{[}5{]}} &
  \multirow{6}{*}{15} &
  GNB &
  \multicolumn{1}{c|}{0.52} &
  \multicolumn{1}{c|}{0.55} &
  \multicolumn{1}{c|}{0.59} &
  \multicolumn{1}{c|}{0.52} &
  0.52 \\ \cline{3-8} 
 &
   &
  RF &
  \multicolumn{1}{c|}{0.59} &
  \multicolumn{1}{c|}{0.55} &
  \multicolumn{1}{c|}{0.55} &
  \multicolumn{1}{c|}{0.72} &
  \textbf{0.67} \\ \cline{3-8} 
 &
   &
  KNN &
  \multicolumn{1}{c|}{0.52} &
  \multicolumn{1}{c|}{0.52} &
  \multicolumn{1}{c|}{0.55} &
  \multicolumn{1}{c|}{0.52} &
  0.55 \\ \cline{3-8} 
 &
   &
  SVM &
  \multicolumn{1}{c|}{0.52} &
  \multicolumn{1}{c|}{0.72} &
  \multicolumn{1}{c|}{0.67} &
  \multicolumn{1}{c|}{0.52} &
  0.52 \\ \cline{3-8} 
 &
   &
  LR &
  \multicolumn{1}{c|}{0.52} &
  \multicolumn{1}{c|}{0.52} &
  \multicolumn{1}{c|}{0.55} &
  \multicolumn{1}{c|}{0.52} &
  0.55 \\ \cline{3-8} 
 &
   &
  XGB &
  \multicolumn{1}{c|}{0.55} &
  \multicolumn{1}{c|}{\textbf{0.72}} &
  \multicolumn{1}{c|}{0.55} &
  \multicolumn{1}{c|}{0.72} &
  0.64 \\ \hline
\textbf{TW} &
  \textbf{15} &
  \textbf{FINDFF} &
  \multicolumn{1}{c|}{\textbf{0.59}} &
  \multicolumn{1}{c|}{0.70} &
  \multicolumn{1}{c|}{\textbf{0.67}} &
  \multicolumn{1}{c|}{\textbf{0.72}} &
  0.64 \\ \hline \hline
{[}8{]} &
  14 &
  NBAME &
  \multicolumn{1}{c|}{0.52} &
  \multicolumn{1}{c|}{0.55} &
  \multicolumn{1}{c|}{0.52} &
  \multicolumn{1}{c|}{0.52} &
  0.55 \\ \hline
\textbf{TW} &
  \textbf{14} &
  \textbf{FINDFF} &
  \multicolumn{1}{c|}{\textbf{0.61}} &
  \multicolumn{1}{c|}{\textbf{0.66}} &
  \multicolumn{1}{c|}{\textbf{0.66}} &
  \multicolumn{1}{c|}{\textbf{0.59}} &
  \textbf{0.64} \\ \hline \hline
\multirow{4}{*}{{[}6{]}} &
  \multirow{4}{*}{6} &
  ELM &
  \multicolumn{1}{c|}{0.52} &
  \multicolumn{1}{c|}{0.55} &
  \multicolumn{1}{c|}{0.55} &
  \multicolumn{1}{c|}{0.52} &
  0.55 \\ \cline{3-8} 
 &
   &
  KNN &
  \multicolumn{1}{c|}{0.52} &
  \multicolumn{1}{c|}{0.67} &
  \multicolumn{1}{c|}{0.55} &
  \multicolumn{1}{c|}{0.59} &
  0.67 \\ \cline{3-8} 
 &
   &
  XGB &
  \multicolumn{1}{c|}{0.52} &
  \multicolumn{1}{c|}{0.67} &
  \multicolumn{1}{c|}{0.60} &
  \multicolumn{1}{c|}{0.52} &
  0.52 \\ \cline{3-8} 
 &
   &
  MARS &
  \multicolumn{1}{c|}{\textbf{0.74}} &
  \multicolumn{1}{c|}{0.59} &
  \multicolumn{1}{c|}{\textbf{0.74}} &
  \multicolumn{1}{c|}{0.62} &
  0.68 \\ \hline
\textbf{TW} &
  \textbf{6} &
  \textbf{FINDFF} &
  \multicolumn{1}{c|}{0.71} &
  \multicolumn{1}{c|}{\textbf{0.71}} &
  \multicolumn{1}{c|}{0.71} &
  \multicolumn{1}{c|}{\textbf{0.62}} &
  \textbf{0.71} \\ \hline \hline
{[}7{]} &
  20 &
  ANN &
  \multicolumn{1}{c|}{0.59} &
  \multicolumn{1}{c|}{0.52} &
  \multicolumn{1}{c|}{0.55} &
  \multicolumn{1}{c|}{0.55} &
  0.52 \\ \hline
\textbf{TW} &
  \textbf{20} &
  \textbf{FINDFF} &
  \multicolumn{1}{c|}{\textbf{0.62}} &
  \multicolumn{1}{c|}{\textbf{0.67}} &
  \multicolumn{1}{c|}{\textbf{0.62}} &
  \multicolumn{1}{c|}{\textbf{0.67}} &
  \textbf{0.71} \\ \hline
\end{tabular}
\caption{\label{tab:iribf}A performance comparison between FINDFF and other previously-developed machine learning models trained on five years of NBA (from 17-18 to 21-22) on the 2020-2021 Iranian Super League Playoffs (Ref: Reference, FC: Feature Count, Alg: Algorithm, TW: This Work)}
\end{table}
\egroup

\subsection{Experiment \RNum{2}}
The purpose of this experiment is to examine the generalizability of methodologies from the first experiment. As we mentioned in Experiment 1, each model is trained and tested on five different years of NBA data. In this experiment, we still train these models on the five NBA datasets but test them on Iranian Super League playoffs. This allows us to compare how generalized each method is when predicting test cases from a significantly different data source. 

\noindent \textbf{Dataset:} For training purposes, we used the same NBA datasets discussed in Experiment 1. But, to test them, we used the 2020-2021 Iranian Basketball Super League playoffs.

\noindent \textbf{Results:} As shown in Table \ref{tab:iribf}, FINDFF models outperformed almost all other methodologies in predicting the outcome of the Iranian Basketball Super League playoffs by a range of 0.02 to 0.12 AUC.

\subsection{Experiment \RNum{3}}
The first two experiments showed how FINs provided higher, and more generalizable performance in  playoff outcome prediction compared to the baselines. After developing MambaNet by building on top of the FINDFF architecture, we aim to demonstrate how integrating other components may affect our hybrid model's performance.

\noindent \textbf{Dataset:} We used the same NBA datasets introduced in Experiment 1. 

\noindent \textbf{Results:} In Table \ref{tab:mamba}, we present the results of our incremental experiment. The first row reports the simplest version of MambaNet using 35 team features that are passed to a FINDFF network imitating the mean (\textit{m}) as a feature. Compared with the baseline, we use a more extensive set of basketball game statistics to form the feature vector of a team since this helps better satisfy the data-intensive requirement of neural networks. At this stage, the AUC varies between 0.70 to 0.72. Next, we trained three more FINs to imitate Standard Deviation \textit{std}, Variance \textit{v}, and Skewness \textit{s} using the same neural network architecture as the mean FIN. The second row of the table shows how adding new signal feature representations improves the AUC up to 0.10 and 0.03 in the 2018-2019 and 2020-2021 NBA datasets. Furthermore, we integrated players' statistics alongside team statistics, leading to an 0.02 increase in AUC across four NBA datasets in the third row. Lastly, as shown in the fourth row, we used RNN layers to create a time-series representation of the game and individual statistics, resulting in 0.03 and 0.02 improvements in 2019-2020 and 2021-2022, respectively.

\bgroup
\def\arraystretch{0.7}
\begin{table}[t]
\centering
\footnotesize
\captionsetup{font=footnotesize}
\begin{tabular}{|c|c|c|c|c|ccccc}
\hline
\multirow{2}{*}{R} &
  \multirow{2}{*}{FS} &
  \multirow{2}{*}{FC} &
  \multirow{2}{*}{IF} &
  \multirow{2}{*}{Layers} &
  \multicolumn{5}{c|}{AUC} \\ \cline{6-10} 
  & 
  & 
  & 
  & 
  &
  \multicolumn{5}{c|}{17-18 \thickspace\thickspace 18-19 \thickspace\thickspace 19-20 \thickspace\thickspace 20-21 \thickspace\thickspace\thickspace 21-22}
   \\ \hline \hline
1 &
  \textit{T} &
  35 &
  \textit{m} &
  Dense &
  \multicolumn{1}{c|}{0.71} &
  \multicolumn{1}{c|}{0.70} &
  \multicolumn{1}{c|}{0.71} &
  \multicolumn{1}{c|}{0.72} &
  \multicolumn{1}{c|}{0.70} \\ \hline
2 &
  \textit{T} &
  35 &
  \textit{\begin{tabular}[c]{@{}c@{}}m\\ std\\ v\\ s\end{tabular}} &
  \begin{tabular}[c]{@{}c@{}}Dense\\ Conv\end{tabular} &
  \multicolumn{1}{c|}{0.71} &
  \multicolumn{1}{c|}{0.80} &
  \multicolumn{1}{c|}{0.71} &
  \multicolumn{1}{c|}{0.75} &
  \multicolumn{1}{c|}{0.70} \\ \hline
3 &
  \textit{\begin{tabular}[c]{@{}c@{}}T\\ \\ P\end{tabular}} &
  \begin{tabular}[c]{@{}c@{}}35\\ \\ 34\end{tabular} &
  \textit{\begin{tabular}[c]{@{}c@{}}m\\ std\\ v\\ s\end{tabular}} &
  \begin{tabular}[c]{@{}c@{}}Dense\\ Conv\end{tabular} &
  \multicolumn{1}{c|}{0.73} &
  \multicolumn{1}{c|}{0.82} &
  \multicolumn{1}{c|}{0.73} &
  \multicolumn{1}{c|}{0.75} &
  \multicolumn{1}{c|}{0.72} \\ \hline
4 &
  \textit{\textbf{\begin{tabular}[c]{@{}c@{}}T\\ \\ P\end{tabular}}} &
  \textbf{\begin{tabular}[c]{@{}c@{}}35\\ \\ 34\end{tabular}} &
  \textit{\textbf{\begin{tabular}[c]{@{}c@{}}m\\ std\\ v\\ s\end{tabular}}} &
  \textbf{\begin{tabular}[c]{@{}c@{}}Dense\\ Conv\\ RNN\end{tabular}} &
  \multicolumn{1}{c|}{\textbf{0.73}} &
  \multicolumn{1}{c|}{\textbf{0.82}} &
  \multicolumn{1}{c|}{\textbf{0.76}} &
  \multicolumn{1}{c|}{\textbf{0.75}} &
  \multicolumn{1}{c|}{\textbf{0.74}} \\ \hline
\end{tabular}
\caption{\label{tab:mamba} Comparing the performance of different MambaNet versions in five years of NBA Playoffs from the
2017-2018 season (17-18) to the 21-22 season. (R: Row, FS: Feature Source, FC: Feature Count, IF: Imitating Feature)}
\end{table}
\egroup

\section{Conclusion}
\label{sec:majhead}
In this work, we tackled playoff basketball game outcome prediction from a signal processing standpoint. We introduced MambaNet, which incorporated historical player and team statistics and represented them through signal feature imitation using FINs. To compare our method with the baseline, we used NBA and Iranian Super League data which enabled us to demonstrate the performance and generalizability of our method. Future studies will potentially use fusion techniques or other suitable data modeling techniques, such as graphs, to develop more advanced neural networks that integrate team and player representations more efficiently to predict playoff outcomes more accurately.

\bibliographystyle{IEEEbib}
\bibliography{strings,refs}

\end{document}